\documentclass{article}

\PassOptionsToPackage{numbers, compress}{natbib}

\usepackage[preprint]{neurips_2024}

\usepackage[utf8]{inputenc} 
\usepackage[T1]{fontenc}    
\usepackage{hyperref}       
\usepackage{url}            
\usepackage{booktabs}       
\usepackage{amsfonts}       
\usepackage{nicefrac}       
\usepackage{microtype}      
\usepackage{xcolor}         
\usepackage{graphicx} %
\usepackage{wrapfig} %
\usepackage{svg}
\usepackage{amsmath,amsfonts, mathtools, bm, bbm, nicefrac}

\DeclareMathAlphabet{\mathsfit}{\encodingdefault}{\sfdefault}{m}{sl}
\SetMathAlphabet{\mathsfit}{bold}{\encodingdefault}{\sfdefault}{bx}{n}

\hypersetup{
    colorlinks=true,
    linkcolor=red,
    citecolor=cyan,
    filecolor=magenta,      
    urlcolor=magenta,
    }

\usepackage[misc]{ifsym}

\newcommand{\Ours}{Loong }
\newcommand{\ours}{Loong}

\title{Loong: Generating Minute-level Long Videos\\ with Autoregressive Language Models}

\author{
  Yuqing Wang\textsuperscript{1} 
  \quad Tianwei Xiong\textsuperscript{1} 
  \quad \bf{Daquan Zhou}\textsuperscript{2}\thanks{Corresponding Author}
  \quad Zhijie Lin\textsuperscript{2} \\ \\
  \quad  \bf{Yang Zhao}\textsuperscript{2} 
  \quad \bf{Bingyi Kang}\textsuperscript{2} 
  \quad \bf{Jiashi Feng}\textsuperscript{2}  
  \quad \bf{Xihui Liu}\textsuperscript{1}\footnotemark[1]  \\ \\
\textsuperscript{1}University of Hong Kong  \quad
\textsuperscript{2}ByteDance \\
}

\begin{document}

\vspace{-30pt}
\maketitle
\vspace{-30pt}
\begin{figure}[htbp]
    \centering
    \includegraphics[width=0.92\textwidth]{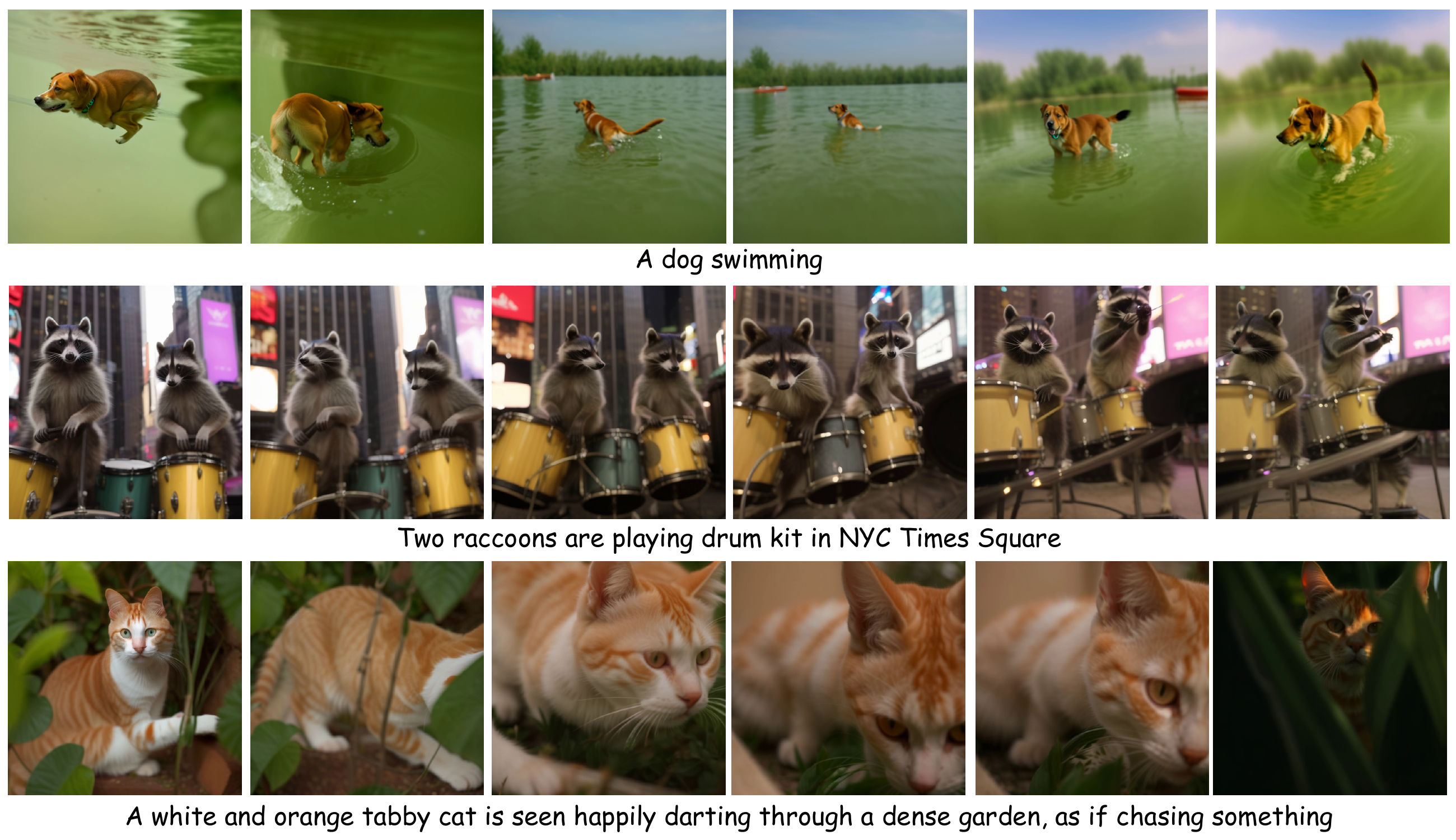}
    \vskip -0.1in
    \begin{minipage}{0.9\textwidth}  
        \caption{\textbf{One-Minute Videos Generated by Loong Conditioned on Texts.} 
        \Ours is an autoregressive LLM-based model that can generate minute-level long videos with consistent appearance, large motion dynamics, and natural scene transitions.}
    \end{minipage}
    \label{fig:vis1}
    \vspace{-3pt}
\end{figure}

\begin{abstract}
\vspace{-2pt}
It is desirable but challenging to generate content-rich long videos in the scale of minutes. Autoregressive large language models (LLMs) have achieved great success in generating coherent and long sequences of tokens in the domain of natural language processing, while the exploration of autoregressive LLMs for video generation is limited to generating short videos of several seconds. In this work, we conduct a deep analysis of the challenges that prevent autoregressive LLM-based video generators from generating long videos. Based on the observations and analysis, we propose \ours, a new autoregressive LLM-based video generator that can generate minute-long videos. Specifically, we model the text tokens and video tokens as a unified sequence for autoregressive LLMs and train the model from scratch. We propose progressive short-to-long training with a loss re-weighting scheme to mitigate the loss imbalance problem for long video training. We further investigate inference strategies, including video token re-encoding and sampling strategies, to diminish error accumulation during inference. Our proposed \Ours can be trained on 10-second videos and be extended to generate minute-level long videos conditioned on text prompts, as demonstrated by the results. More samples are available at: \url{https://yuqingwang1029.github.io/Loong-video}.
\end{abstract}

\clearpage
\section{Introduction}

\vspace{-10pt}

Over the past few years, video generation models, including diffusion-based ones~\cite{video-diffusion-models,gen1_paper,make-a-video,bar2024lumiere,MakePixelsDance,girdhar2023emu,zhou2022magicvideo,wang2024magicvideo} and language model based approaches~\cite{yan2021videogpt,kondratyuk2023videopoet}, have shown impressive results in generating short videos of a few seconds.
To capture more comprehensive content, it is desirable to generate long videos with consistent appearance, larger motion dynamics, and natural scene transitions.
Despite recent works~\cite{wang2023gen_L,streamingt2v,openai2024sora} to generate long videos with diffusion-based video generators, generating content-rich long videos on the scale of minutes remains largely underexplored and challenging.

Autoregressive large language models (LLMs) have shown remarkable success in generating long and coherent text sequences~\cite{radford2018improving, radford2019language, gpt3, reid2024gemini, touvron2023llama, touvron2023llama2}, demonstrating their ability to capture long-range dependencies and complex temporal patterns. 
Inspired by the success of autoregressive LLMs in other modalities and their flexibility in unifying various modalities and tasks, recent works~\cite{yan2021videogpt,kondratyuk2023videopoet} have explored autoregressive language models for video generation.
Those approaches map videos into discrete tokens and use text tokens as conditioning to generate the video tokens by next-token prediction with decoder-only transformers.
State-of-the-art autoregressive LLM-based video generator~\cite{kondratyuk2023videopoet} can generate high-quality 2-second short video clips and iteratively extend to 10-second coherent videos.

Despite demonstrating the ability of long sequence generation in NLP and being explored for short video generation, the potential of LLMs to generate minute-level, content-rich, and dynamic videos remains unexplored. In natural language processing, LLMs can be trained on long sequences and extended beyond the training length. However, we empirically observe that either training autoregressive LLMs on long video sequences or extending short video generators to generate long videos leads to unsatisfactory performance for minute-level video generation. A question arises: \textbf{\textit{What restricts the capability of autoregressive language models for generating long videos?}}

We hypothesize that the main obstacles are the large redundancy and strong inter-frame dependency among video tokens.
The video tokens of the current frame depend heavily on the tokens of the previous frames, leading to two challenges for long video generation:
(1) \textbf{\textit{Imbalanced loss during training.}} When trained with the next-token prediction objective, predicting early-frame tokens from text prompts is much more difficult than predicting late-frame tokens based on the ground-truth tokens of previous frames. The imbalanced difficulty levels of tokens lead to imbalanced loss during training. The issue becomes more severe as the video length increases, where the accumulated loss of many easy tokens largely surpasses the loss of a few difficult tokens and dominates the gradient direction.
(2) \textbf{\textit{Error accumulation during inference.}} While the model predicts the next token conditioned on previous \textit{ground-truth} tokens during training, it has to predict the next token conditioned on previous \textit{predicted} tokens during inference. This training-inference discrepancy leads to error accumulation during inference. Because of the strong inter-frame dependency among video tokens and the large number of video tokens, such error accumulation is non-negligible and causes visual quality degradation for long video inference.

In this work, we propose \textbf{\ours}, aiming to unleash the power of autoregressive language models to generate long videos in the scale of minutes. 
Our autoregressive LLM-based video generator consists of two components: a video tokenizer that compresses videos into sequences of discrete video tokens, and an autoregressive LLM that models the unified sequence of text tokens followed by video tokens through next-token prediction.
To mitigate the problem of imbalanced loss for long video training, we introduce a progressive short-to-long training strategy that gradually increases the training video length. We further propose loss re-weighting for early frames to prevent the model from being dominated by many easy tokens in the late frames. Moreover, we investigate inference strategies, including the video token re-encoding and sampling strategy, to further extend the video length by iteratively generating the next frames conditioned on previously generated frames. In order to enable training and inference with longer videos, we adopt low-resolution videos for the LLM-based video generator, and leverage a super-resolution and refinement module to further enhance the resolution and fine-grained details of the generated long videos.

In summary, we propose \ours, a novel autoregressive LLM-based video generator that can generate content-rich, coherent, and dynamic long videos in the scale of minutes. Based on our observations and analysis of the issues that limit the power of LLMs for long video generation, we propose progressive short-to-long training with a loss weighting scheme to enable model training on 10-second videos. We further investigate inference strategies to extend the 10-second videos to minute-level videos by autoregressive generation strategies designed for long video inference. Our model demonstrates its ability in generating minute-level long videos through extensive experiments.

\section{Related Work}
\label{related_work}
\textbf{Video generation.}
The mainstream video generation methods can be categorized into GAN-based~\cite{goodfellow2014generative, vgan, mocogan}, Diffusion-based~\cite{ho2020ddpm, zhou2022magicvideo,ho2022video, ho2022imagen,SVD, make-a-video,girdhar2023emu, li2023videogen, zhang2023show1,openai2024sora} and language-model-based~\cite{vaswani2017attention,kondratyuk2023videopoet,villegas2022phenaki,yu2023language}. Among them, Diffusion-based methods have recently gained the most popularity. Most Diffusion-based methods encode videos into latent space~\cite{rombach2022high} for efficient training and utilize progressive inference strategies~\cite{ho2022imagen, he2022lvdm, AlignYourLatents} to generate videos with high spatial-temporal resolution. With a new scalable Diffusion Transformer~\cite{peebles2023scalable} architecture, Sora~\cite{openai2024sora} has further pushed video generation to a new stage. Different from diffusion-based video generation methods, our work aims to explore and unleash the potentiality of language models for long video generation, as their ability for modeling long sequence and scaling up have been proved in NLP.

\textbf{Image and video generation with language models.}
Language models have recently been explored for visual generation, with most works focusing on tokenizing visual data into a form that can be processed by these models. Quantization techniques like VQ-VAE \cite{vqvae,esser2020taming} are commonly used, and transformers are employed to model the resulting tokens.
For image generation, autoregressive or masked transformers are prevalent \cite{ramesh2021dalle, yu2022scaling, chang2023muse, chang2022maskgit, yu2024spae, sun2024autoregressive, tian2024visual}. In short video generation, image-level or video-level tokenizers are utilized, incorporating spatial-temporal compression and causal structures. Transformers model the spatial-temporal relationships, with various techniques proposed, such as sparse attention, spatial-temporal attention, large-scale pre-training, and improved tokenization \cite{yan2021videogpt,ge2022long, wu2021godiva, hong2022cogvideo, yu2023magvit}. VideoPoet \cite{kondratyuk2023videopoet} stands out as a multimodal model using bidirectional attention for conditioning, while our method aligns better with the language model paradigm by using unidirectional attention for both text and video.
However, these short video generation models focus on producing 1-5 second clips, limiting their ability to capture complex events and maintain consistency over longer durations.

\textbf{Long video generation.}
Previous works have explored long video generation using various approaches. LongVideoGAN~\cite{brooks2022generating}, NUWA-XL~\cite{yin2023nuwa}, and GAIA-1~\cite{hu2023gaia} utilized GAN-based methods, diffusion-over-diffusion techniques, or world models but were limited to specific domains. More recently, video diffusion models have been extended for longer video generation. FreeNoise~\cite{qiu2023freenoise} and Gen-L~\cite{wang2023gen_L} focus on sampling noise vectors and aggregating overlapping short video segments, respectively, while StreamingT2V~\cite{streamingt2v} proposes an autoregressive approach with memory blocks for consistency and appearance preservation. In the language model domain, Phenaki~\cite{villegas2022phenaki} generates variable-length videos using a masked video transformer. Despite these advancements, generating long videos with rich motion dynamics, consistent appearance, and high visual quality in the open domain remains a challenge.
\section{Method}
\label{method}

We present \ours, an autoregressive LLM-based model for generating long videos in the scale of minutes.
We introduce the overall framework, composed of the video tokenizer and the LLM-based video generator, in Sec.~\ref{sec:3.1}. We analyze the problem with long video training and propose the progressive short-to-long training with loss re-weighting scheme, enabling training on 10-second videos, in Sec.~\ref{sec:3.2}. We further investigate inference strategies to extend the generated video length to the minute level and post-processing techniques to enhance the spatial resolution of generated videos in Sec.~\ref{sec:3.3}.

\subsection{Overall Framework}
\label{sec:3.1}

\begin{figure}[htbp]
    \centering
    \includegraphics[width=0.85\textwidth]{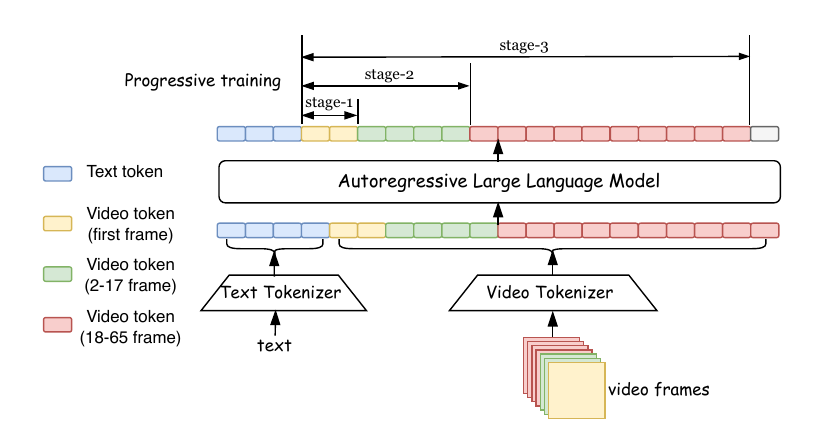}
    \vskip -0.1in
    \caption{\textbf{Overall Framework and the Training process of \Ours.} Given the input text tokens, the model predict video tokens autoregressively. All the text and video information is formulated into a unidirectional discrete token sequence, where the model predicts the next token based on the previous tokens. Video Tokenizer is utlized to convert video frames into discrete tokens. We use different color to represent first frame, short clip and long clip separately. We follow a progressive training pipeline to train on long videos. We omit the special tokens for simplicity.}
    \vskip -0.1in
    \label{fig:stage1}
\end{figure}


Inspired by previous work in LLM-based image generation and video generation models~\cite{ramesh2021dalle,chang2022maskgit,yu2023magvit,yu2023language,kondratyuk2023videopoet}, \Ours is designed with two components: a video tokenizer that efficiently compresses the videos into discrete tokens, and a decoder-only transformer that autoregressively predicts next video tokens based on text tokens.


\textbf{Video Tokenizer.}
In order to enable spatial-temporal joint compression and joint modeling of images and videos, we leverage causal 3D CNN architecture for the tokenizer, inspired by MAGViT2~\cite{yu2023language}. 
The encoded spatial-temporal features are quantized into discrete tokens with Clustering Vector Quantization (CVQ)~\cite{Zheng_2023_CVQ}, an improved version of VQGAN~\cite{esser2020taming} designed to enhance codebook utilization. To extend the temporal coverage of videos within a limited number of tokens, we work with low-resolution videos and leave super-resolution for the post-processing in Sec.~\ref{sec:3.3}. The tokenizer can compress a 10-second video (65 frames, $128 \times 128$ resolution for each frame) into a sequence of $17 \times 16 \times 16$ discrete tokens with a vocabulary size of 8192.

\textbf{Autoregressive LLM-based Video Generation.}
With the video frames converted into discrete tokens, we can now model the text and video tokens as a unified sequence and formulate text-to-video generation as autoregressively predicting video tokens conditioned on the text tokens with decoder-only Transformers. The process is illustrated in Fig.~\ref{fig:stage1}. For simplicity, we omit the special separate tokens in the following formulation.
Let $\mathbf{t} = \{t_1, t_2, \ldots, t_N\}$ represent the sequence of text tokens, where $N$ is the number of text tokens. Similarly, let $\mathbf{v} = \{v_1, v_2, \ldots, v_L\}$ represent the sequence of video tokens, where $L$ is the number of video tokens.
The autoregressive LLM models the unified token sequence $\mathbf{s} = [\mathbf{t}; \mathbf{v}]$ and is trained with the next-token prediction loss for the video tokens.
\begin{equation}
\mathcal{L} = -\sum_{i=1}^{L} \log p(v_i \mid v_{<i}, \mathbf{t})
\end{equation}
where $v_i$ denotes the $i$-th token in the video sequence $\mathbf{v}$, and $v_{<i}$ denotes all the video tokens preceding $v_i$.

\textbf{Discussion.}
Different from 
VideoPoet~\cite{kondratyuk2023videopoet}, which encodes text with a pretrained T5 text encoder~\cite{2020t5} and applies bidirectional attention for the input condition tokens and causal attention for the video tokens, our approach does not rely on a pretrained text encoder. Instead, we formulate the text tokens and video tokens as a unified token sequence and apply causal attention to all tokens.
Our unified autoregressive modeling of text tokens and video tokens provides a simpler formulation that is consistent with modern GPT-style LLMs~\cite{gpt3}. This design may lead to potential benefits in extending our model to multimodal LLMs that unify different modalities and different tasks for understanding and generation.

\subsection{Progressive Short-to-Long Training with Loss Re-weighting}
\label{sec:3.2}

Most video generation models are trained on short video clips, typically no more than 4 seconds, which limits their ability to capture long-term dependencies and complex dynamics in longer videos. To address this limitation, it is essential to train these models on videos with longer durations, enabling them to learn and generate more coherent and contextually rich video content.

\begin{wrapfigure}{r}{0.45\textwidth}
    \vspace{-20pt} 
    \centering
    \includegraphics[width=\linewidth]{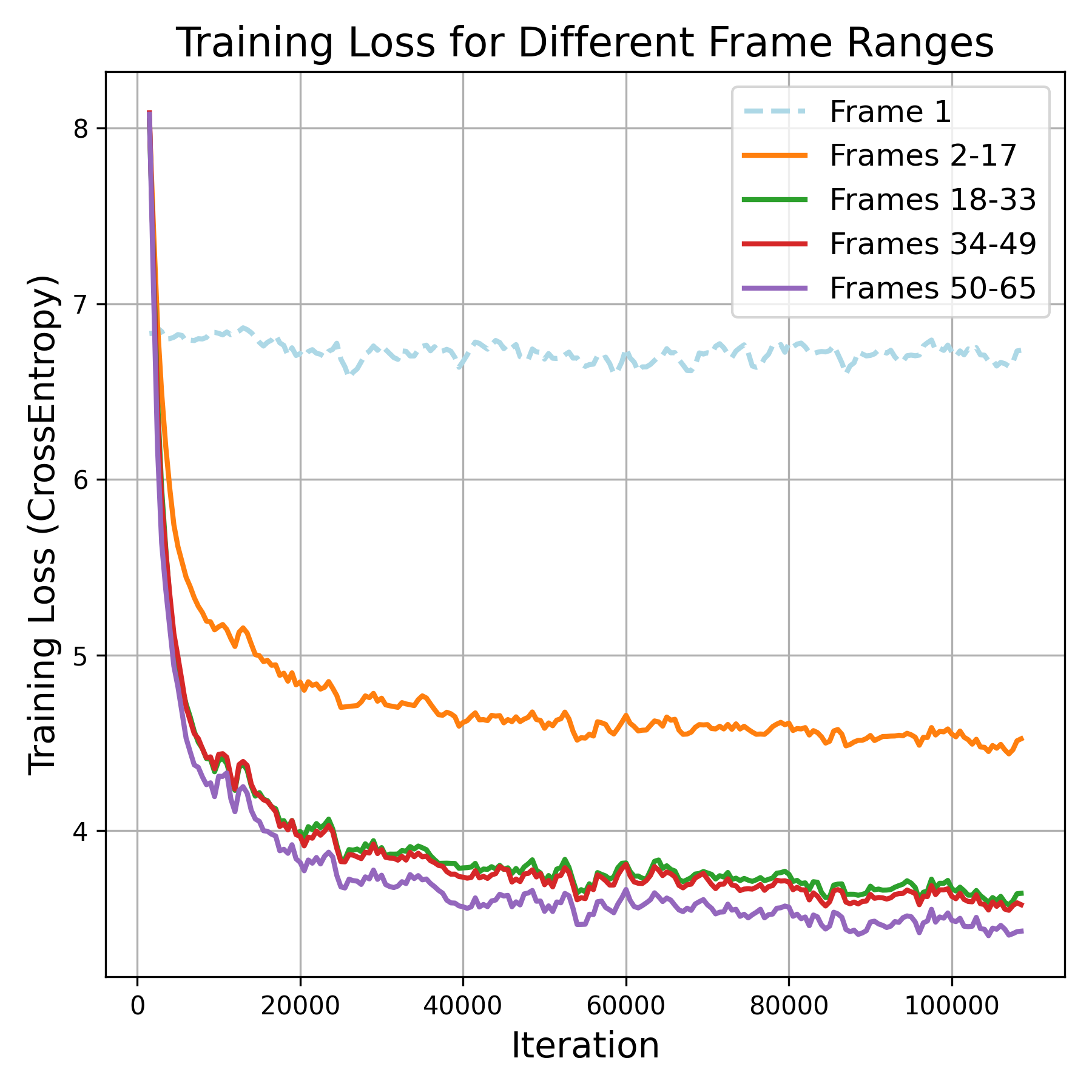}
    \vskip -0.15in
    \caption{\textbf{Imbalanced Training Losses When Training Directly on Long Videos.} The training loss for late frames (18-65) is smaller than that of early frames (1-17), and the loss for the first frame remains relatively high, leading to suboptimal visual quality in the early frames( despite the model being pretrained on text-to-image).
    }
    \label{fig:clip_loss_curve}
    \vskip -0.15in
    \vspace{-6pt} 
\end{wrapfigure}

However, training directly on long videos leads to suboptimal performance, even when the model is trained for a large number of iterations. We illustrate the loss curve of different frame ranges when training on 65-frame videos (with 4,356 tokens, covering 10 seconds) in Fig.~\ref{fig:clip_loss_curve}. We empirically observe that tokens from early frames (frames 1-17) have larger losses than those from later frames (tokens from frames 50-65 have the smallest average loss).
During training, the model learns through next-token prediction, where it is much easier to predict tokens of later frames given the previous ground-truth video and text tokens.
In comparison, predicting early-frame tokens with little visual cues from previous frames is more challenging. 
The imbalanced loss is a severe problem for long-sequence training because the accumulated loss of the many easy-to-predict tokens from later frames (18-65) surpasses the loss of the few difficult-to-predict tokens from early frames (1-17) and dominates the gradient direction, leading to suboptimal visual quality in the generated videos.

To mitigate the aforementioned challenge of imbalanced video token difficulties, we propose a progressive short-to-long training strategy with loss re-weighting, demonstrated in the following.

\textbf{Progressive short-to-long training.}
In order to allow the model to first learn the text-conditioned appearance and motion of short videos, and then smoothly adjust to longer-range dependencies and more complex motion patterns in longer videos, we factorize training into three stages which gradually increases the training video length, as illustrated in the Fig.~\ref{fig:stage1}:
(1) In \textit{stage-1}, we pretrain the model with text-to-image generation on a large dataset of static images, which helps the model to establish a strong foundation for modeling per-frame appearance and structure. (2) In \textit{stage-2}, we continue to train the model jointly on images and short video clips of 17 frames, where the model learns to capture short-term temporal dependencies and motion patterns while preserving the per-frame visual quality. 
(3) In \textit{stage-3}, we increase the number of video frames to 65, covering a temporal range of 10 seconds, and continue joint training. 

\textbf{Loss re-weighting for early frames.}
To further strengthen the supervision of early frames and to prevent the model from forgetting the stage-1 and stage-2 priors, we propose a loss re-weighting scheme for stage-3. To be specific, we apply larger loss weights for the tokens of early frames, and the overall weighted loss is formulated as
\begin{equation}
\label{eq:eq2}
\mathcal{L}_{\text{weighted}} = -(1+\lambda) \sum_{i=1}^{K} \log p(v_i \mid v_{<i}, \mathbf{t}) -\sum_{i=K+1}^{L} \log p(v_i \mid v_{<i}, \mathbf{t}),
\end{equation}
where the first term denotes the loss for the $K$ tokens corresponding to the early frames (the first 17 frames), and the second term denotes the loss for the $L-K$ tokens corresponding to the later frames (frames 18-65). $\lambda$ is a positive value to strengthen the loss weight of early frames.

With the loss weighting and progressive training strategy, our model effectively mitigates the issues of long video training. As the model is trained on a temporal range of 10 seconds, it can generate videos of up to 10 seconds with improved temporal coherence and consistency while maintaining the strong appearance and motion priors learned from the image and short video clips.

\subsection{Inference Strategies for Extending Video Length and Resolution}
\label{sec:3.3}

\begin{figure}[htbp]
    \centering
    \includegraphics[width=0.65\textwidth]{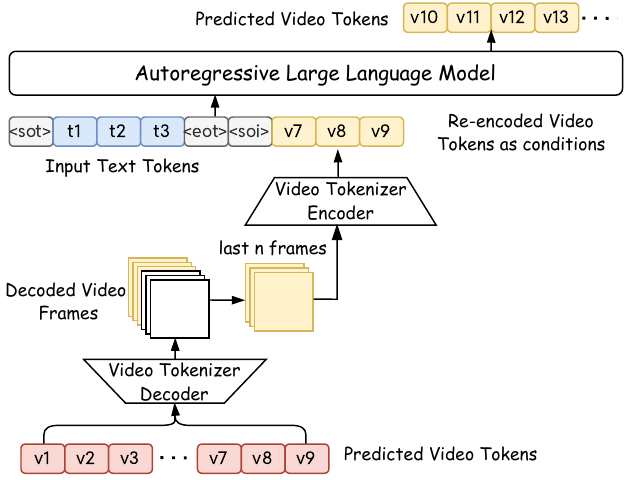}
    \vskip -0.1in
    \caption{\textbf{Inference process of \Ours.} Given the input text, the model first predicts video tokens (illustrated by \texttt{v1}-\texttt{v9}) for the first 10s. The tokens from the last n frames of this clip are then decoded into video frames and re-encoded by the video tokenizer. These re-encoded tokens (\texttt{v7}-\texttt{v9}), along with the text tokens, serve as conditions to predict the video tokens (\texttt{v10}-\texttt{v13}) for the next clip. This iterative process of token prediction, partial decoding, and re-encoding enables extending videos beyond the training duration while mitigating quality degradation. This process is repeated until the generated video reaches the desired length.}
    \vskip -0.2in
    \label{fig:stage2}
\end{figure}

Large language models are proven to be length-generalizable, so we expect the LLM-based video generator trained on 10-second videos to be extended to generate longer videos autoregressively.
However, generalizing beyond the training video duration is non-trivial and may lead to error accumulation and quality degradation. 
For instance, a one-minute video corresponds to approximately $26,112$ video tokens under our current settings, which is significantly longer than most text sequences typically encountered in language modeling tasks. 
The considerable length and the large inter-frame dependency among video tokens pose challenges for extending the LLM-based generator for long video generation.
In this subsection, we investigate inference strategies to generate minute-level videos and post-processing methods like video super-resolution and refinement to generate higher-quality videos.

\textbf{Video token re-encoding.}
A natural way of extending videos beyond the training duration is to iteratively generate the tokens of the next video clip, conditioned on the text prompts and the previously generated tokens of the current video clip, exploiting the benefit of autoregressive language models.
However, this strategy leads to severe video quality degradation for video frames beyond the training range.
With further analysis, we find that this issue stems from the token misalignment caused by the causal video tokenizer. To be specific, the tokens from the last $n$ frames in a video clip are derived based on the context of all previous frames, while the tokens from the first $n$ frames in a new video clip are derived without the context of the previous video clip. Therefore, generating tokens for the new clip directly conditioned on previous tokens leads to distribution shift in the input features for LLMs. To address this issue, we decode the LLM-generated video tokens to the pixel-space videos and then re-encode the last $n$ frames with the video tokenizer. The re-encoded video tokens and the text tokens serve as the conditions to generate the tokens of the next video clip. The inference process is illustrated in Fig.~\ref{fig:stage2}.

\textbf{Sampling strategy.}
Decoding video tokens with autoregressive language models is prone to \textit{error accumulation} because of the autoregressive nature of the model and the strong inter-frame dependencies of video tokens.
Errors in predicting one token can propagate and influence the generation of subsequent tokens, leading to a degradation in video quality as the length increases.
To mitigate this issue, we draw inspiration from the Top-$k$ sampling strategy commonly used in NLP tasks. During the token sampling process, we only sample from the Top-$k$ most probable tokens, ensuring that the generated tokens are of high quality. 
By focusing on the most likely tokens, we reduce the influence of potential errors on subsequent token generation, effectively alleviating the error accumulation problem. On the other hand, we also observe that too small values of $k$ ($k=1$ degrades to greedy decoding) lead to almost static videos with little motion. To balance dynamic motion and error accumulation, we choose $k=50$ for our model.

\textbf{Super-resolution and refinement.}
As introduced in Sec.~\ref{sec:3.1}, our video tokenizer and LLM-based video generator operates on the low-resolution $128\times 128$ videos. This design trades off spatial resolution for longer video sequences during training and inference. We apply off-the-shelf super-resolution and refinement models~\cite{stable_diff,guo2023animatediff,controlnet,mokady2023null} on the LLM-genereated low-resolution videos. This module serves as a post-processing to enhance the spatial resolution and fine-grained visual details of videos, without affecting the main content and motion of the generated videos. 

\section{Experiments}
\label{exp}

\subsection{Implementation Details}
\label{sec:4.1}

\textbf{Model Architecture.} Our video generation model follows the same architecture    as LLaMA~\cite{touvron2023llama}, with the model size ranging from 700M to 7B parameters. We train the models from scratch, without using any text-pretrained weights. The vocabulary consists of 32,000 tokens for text, 8,192 tokens for video, and 10 special tokens, resulting in a total vocabulary size of 40,202. 
For the video tokenizer, we attempt to reproduce the architecture of MAGVIT2~\cite{yu2023language}, which is a causal 3D CNN structure that separately models the first frame of the video. The model compresses the spatial dimensions (width and height) by a factor of 8 and the temporal dimension by a factor of 4.  We utilize the Clustering Vector Quantization (CVQ)~\cite{Zheng_2023_CVQ} method for quantization, as it achieves a higher codebook usage ratio compared to the original Vector Quantization (VQ)\cite{esser2020taming, VQ_VAE_paper} approach. The video tokenizer has a total of 246M parameters. 

\textbf{Training.}
Our models are trained on 100M text-image pairs filtered from the combination of the CC12M~\cite{sharma2018conceptual} and LAION-2B~\cite{schuhmann2022laion} datasets, as well as the WebVid-10M~\cite{Bain21} video training set and 5.5M video clips filterd from HDVG~\cite{videofactory}. The training process follows the progressive strategy described in Sec.~\ref{sec:3.2}. We first pre-train the model on the combined image dataset for 200k iterations, followed by joint training on images and 17-frame video clips from the combined video dataset for another 200k iterations with a batch size of 512. We then jointly train on 65 frames (covering 10 seconds) for 100k iterations with a batch size of 256. The $\lambda$ is set to 1.0 for the weighted loss of Eq.~\eqref{eq:eq2}. In each stage, we use AdamW optimizer with a base learning rate of 1.0e-4. The learning rate is scheduled using a linear warmup for the first 10,000 iterations, followed by a cosine annealing decay until reaching the maximum iteration count. For the training of the tokenizer, we also use a progressive approach on the same dataset, increasing the video length from 1 to 17 to 65 frames while maintaining a resolution of $128 \times 128$, with a batch size of 64 and training for 400k iterations.

\textbf{Inference.}
During inference, our model first generates the initial 65 frames based on the text prompt. We then use the last 5 predicted frames as conditions for video extension. The classifier-free guidance ratio is set to 7.5.

\subsection{Ablation Study}
In this section, we conduct ablation studies to evaluate the effectiveness of our main design choices. Unless otherwise specified, we use the 3B model with an output spatial resolution of $128\times 128$, without any super-resolution  and refinement module. To reduce computational cost, we train the models for half the number of iterations compared to the full setting described in Sec.~\ref{sec:4.1}. Due to the lack of a general long video generation benchmark, we build a custom one by selecting the top-1000 longest clips from the WebVid~\cite{Bain21} validation set and slicing each to 27 seconds, the duration of the shortest among them. We employ two commonly used video generation metrics on this benchmark: Fréchet Video Distance (FVD)\cite{unterthiner2018towards} and Video-Text Matching (VTM) score calculated by CLIP (ViT-L/14)\cite{CLIP_paper}. 
We use the text prompt sets from prior works~\cite{bar2024lumiere,girdhar2023emu,imagen_video,make-a-video,openai2024sora} to generate videos for visualization. 

\textbf{Model Scaling.}
Scalability is an important characteristic of LLMs. To study scaling behavior of our model, we evaluate performance of the models with different sizes. Tab.~\ref{tab:scaling} presents the quantitative results of our models with 700M, 3B and 7B parameters using the same number of iterations on the custom benchmark. We observe that larger models achieve better FVD and VTM scores, demonstrating the scalability of model size for our approach.  

\begin{wraptable}{r}{0.35\textwidth}
    \vskip -0.2in
    \centering
    \caption{\textbf{Model Size Scalability of \Ours.} 
    The performance improves as the model size increases.
    }
    \begin{tabular}{@{}lcc@{}}
    \toprule
     & $\mathrm{FVD}_{\mathrm{I3D}}$↓  &$\mathrm{VTM}_{\mathrm{c}}$↑\\
    \midrule
    700M & 633 & 21.5 \\
    3B & 572 & 22.8 \\
    7B & \textbf{432} & \textbf{24.1} \\
    \bottomrule
    \end{tabular}
    \label{tab:scaling}
\end{wraptable}

\begin{figure}[t]
    \centering
    \includegraphics[width=1.0\textwidth]{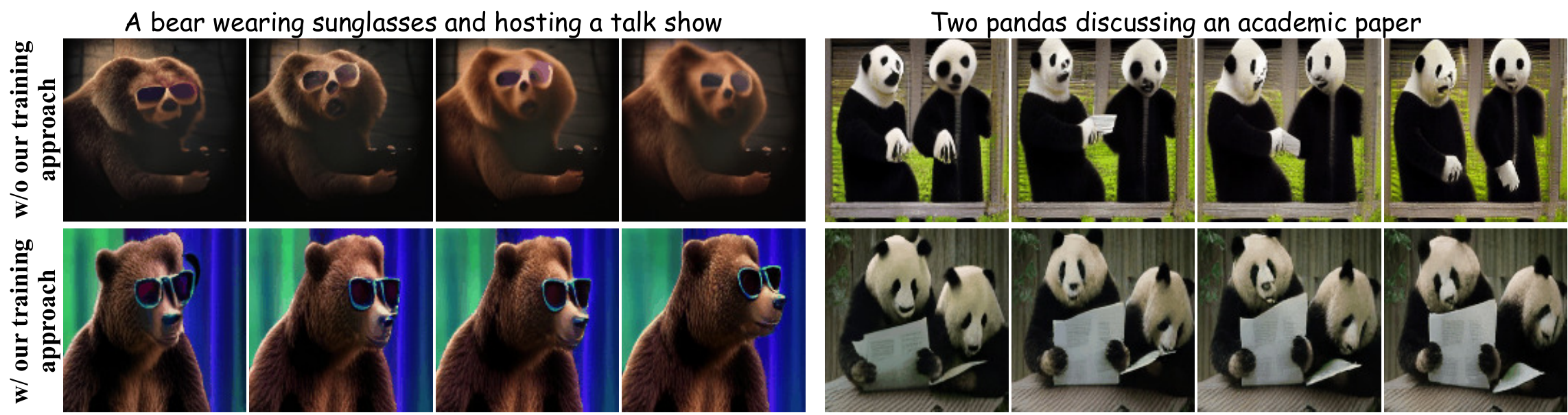}
    \vskip -0.1in
    \caption{\textbf{Effectiveness of the Progressive Training with Loss Re-weighting.} We sample 4 frames from the 17 earlier frames of the video generation results, to show the performance of models trained with or without our training strategy. The top row shows results of the model trained directly on long video, the appearance of objects degrades largely. The bottom row shows the results model trained with our proposed training approach, the appearance preserves effectively. 
    }
    \vskip -0.15in
    \label{fig:train}
\end{figure}

\textbf{Progressive Training with Loss Re-weighting.}
To validate the effectiveness of our proposed training strategy, we compare the models trained with and without our proposed strategies. Both models are pre-trained on images and then trained on videos. Fig.~\ref{fig:train} (top row) shows the generated frames of model trained by a single training stage without our proposed strategy. It is clear that the videos generated by the directly-trained models suffer from significant object appearance degradation, losing much of the structure information. In contrast, videos generated by the model trained with the proposed approach effectively preserve the appearance details.

\begin{figure}[t]
    \centering
    \includegraphics[width=1.0\textwidth]{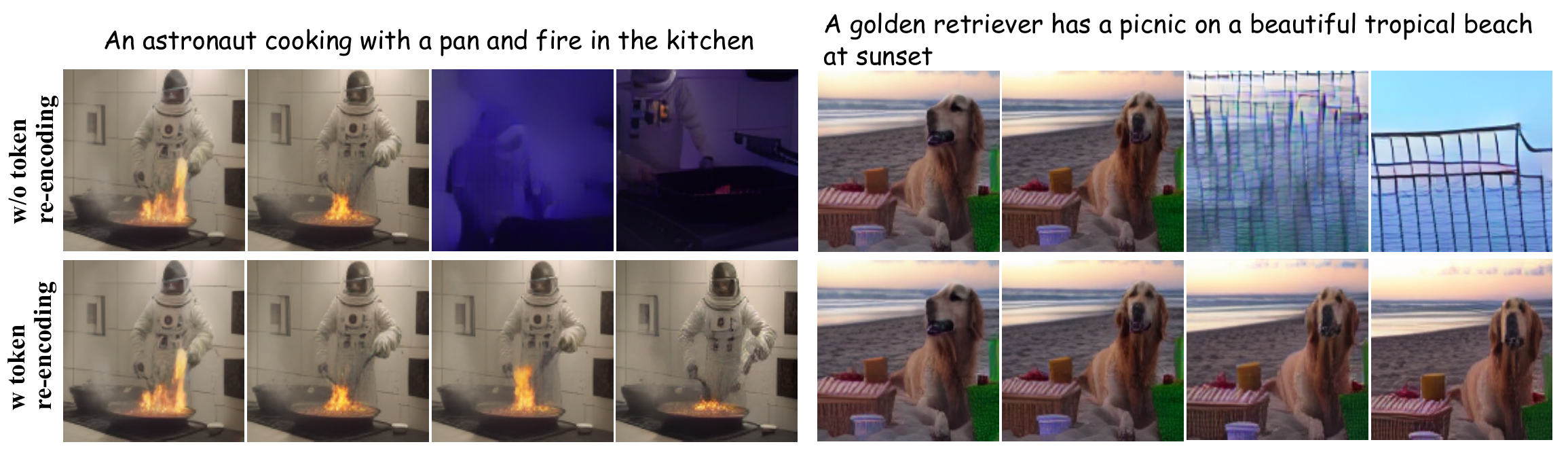}
    \vskip -0.1in
    \caption{\textbf{Effectiveness of Token Re-encoding during Video Extension.} For each sample, the left two images show the results before the extension process, and the right two images show the results after extension. Without token re-encoding, the extension fails to generate visually consistent content. 
    }
    \vskip -0.2in
    \label{fig:tokenrealign}
\end{figure}

\begin{wrapfigure}{r}{0.45\textwidth} 
    \vskip -0.15in
    \includegraphics[width=\linewidth]{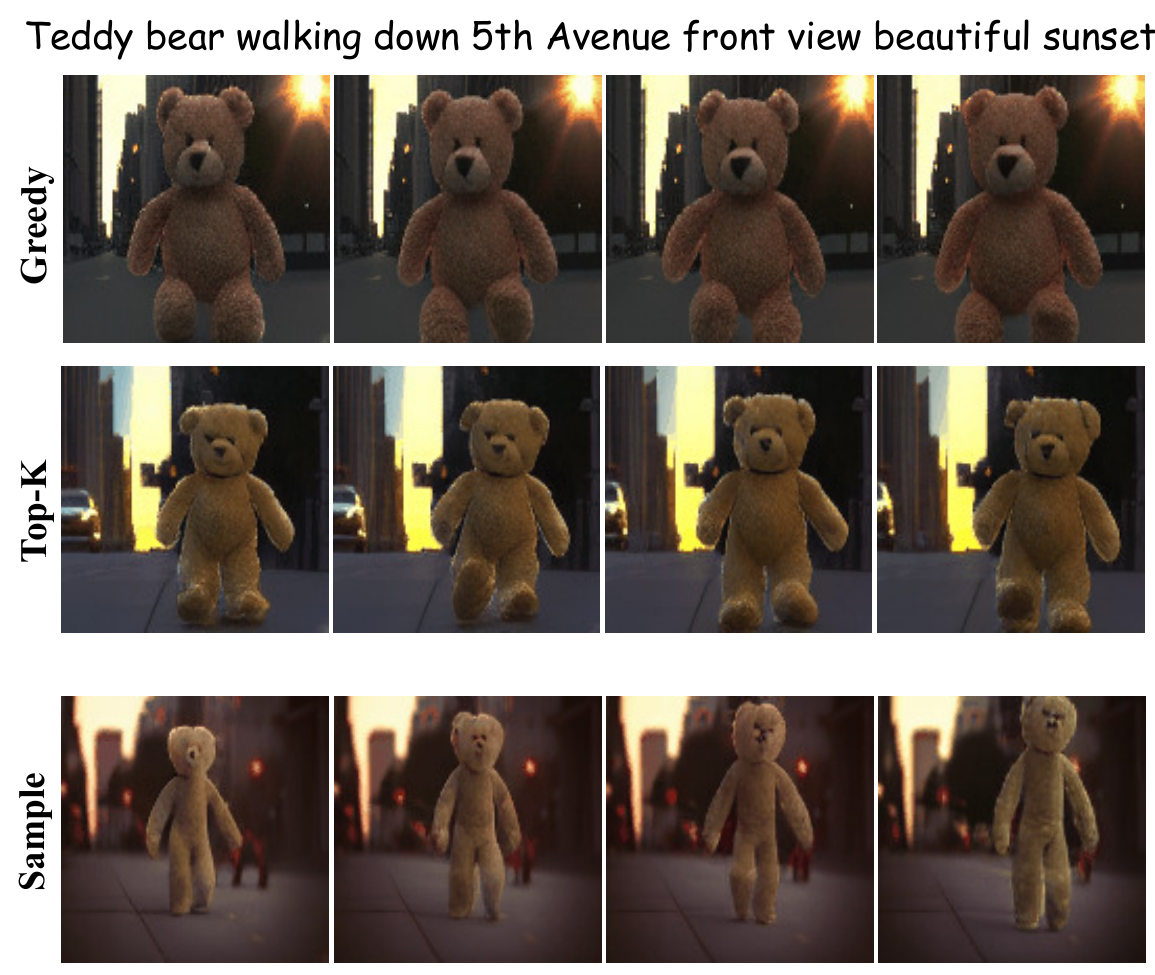}
       \caption{\textbf{Study on Sampling Strategies.} Results of three different inference sampling strategies. Greedy decoding produces stable results but lacks diversity between frames. Multinomial sampling generates more dynamic and diverse content but with lower quality. Top-$k$ sampling achieves a balance between stability and diversity. $k$ is set to 50 in this experiment. 
    }
    \label{fig:sample}

\end{wrapfigure}

\textbf{Video Token Re-encoding.} Fig.~\ref{fig:tokenrealign} illustrates the importance of token re-encoding during the video extension process.  
Without proper token re-encoding, the model fails to maintain visual consistency when extending the video, resulting in abrupt changes in appearance and content. In contrast, by employing our token re-encoding technique, the extended frames seamlessly continue the video with coherent visual style and content.

\textbf{Sampling Strategy for Inference.}
We compare three sampling strategies when predicting each token: greedy decoding ($k=1$), top-$k$ sampling, and multinomial sampling from the whole vocabulary ($k$ equals video token vocabulary size). As shown in Fig.~\ref{fig:sample}, greedy decoding generates stable results but lacks diversity, while multinomial sampling produces more dynamic content at the cost of quality. Top-$k$ sampling ($k=50$) balances stability and diversity. A smaller $k$ value prioritizes stability, resulting in less diverse motion, while a larger $k$ allows for more dynamic and varied content at the risk of introducing instability. In the process of video extension, selecting an appropriate $k$ value is crucial for maintaining consistency and mitigating error accumulation over longer sequences.
\subsection{Quantitative Results}

\begin{table}[h]
    \centering
    \caption{\textbf{Comparison of zero-shot text-to-short-video generation on the MSRVTT benchmark}}
    \label{tab:short_sota}
    \begin{small}
    \resizebox{1.0\textwidth}{!}{ 
    \begin{tabular}{@{}lcccccc@{}}
    \toprule
    Model & CogVideo\cite{hong2022cogvideo} & MagicVideo\cite{zhou2022magicvideo} & ModelScopeT2V\cite{Modelscope} & Show-1\cite{zhang2023show1} & VideoPoet\cite{kondratyuk2023videopoet} & \Ours \\
    \midrule
    CLIPSIM & 0.2631 & - & 0.2930 & 0.3072 & 0.3049 & 0.2903 \\
    FVD & 1294 & 998 & 550 & 538 & 213 & 274 \\
    \bottomrule
    \end{tabular}
    }
    \end{small}
    \vskip -0.1in
\end{table}

\textbf{Zero-shot Text to Short Video Generation.}
Although our approach is not specifically designed for short video generation, we compare our performance on the MSR-VTT dataset~\cite{xu2016msr} using CLIP similarity (CLIPSIM)~\cite{wu2021godiva} and FVD~\cite{unterthiner2018towards} metrics, evaluated on 16 frames. As shown in Tab.~\ref{tab:short_sota}, our FVD score is the second-best, only slightly behind VideoPoet~\cite{kondratyuk2023videopoet} (pretrained). However, our CLIPSIM score is lower compared to some other methods, which can be attributed to the fact that our approach is trained from scratch without utilizing any pre-trained text weights. In contrast, methods with higher CLIPSIM scores, such as VideoPoet, leverage pre-trained language models like T5~\cite{2020t5} for text encoding, while diffusion-based methods often employ CLIP~\cite{CLIP_paper} text embeddings, which are already trained on the CLIP dataset. Despite not using pre-trained text models, our method still achieves competitive performance, demonstrating its effectiveness in capturing the semantic relationship between text and video.

\begin{wrapfigure}{r}{0.35\textwidth} 
   \vskip -0.15in
    \includegraphics[width=\linewidth]{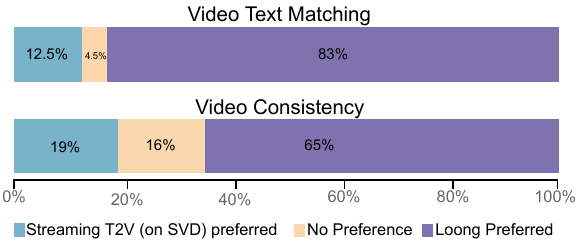}
    \caption{\textbf{User Study on 1-min videos.} Comparison with the StreamingT2V on SVD model. Our model is more preferred by human raters in terms of both visual text match and content consistency. }
    \label{fig:user study}
\end{wrapfigure}
\textbf{User Study on Long Video Generation.} We conduct a user study to compare our method with StreamingT2V~\cite{streamingt2v}, a state-of-the-art open-sourced long video generation method built on Stable Video Diffusion~\cite{SVD}. We use 50 text prompts from prior works~\cite{bar2024lumiere,girdhar2023emu,imagen_video,make-a-video} to generate 1-min videos. In the study, users are presented with 2 videos generated by the two models, conditioned on the same text. They are asked to choose the preferred video based on visual text matching and content consistency. The videos are presented randomly, and users are not informed about the models. We collect 440 responses. As shown in Fig.~\ref{fig:user study}, our model outperforms StreamingT2V in both content consistency (win rate 0.83 vs. 0.125) and visual text matching (win rate 0.65 vs. 0.19).

\subsection{Visualization Results}
\label{sec:exp_visualization}

\begin{figure}[htbp]
    \centering
    \includegraphics[width=1.0\textwidth]{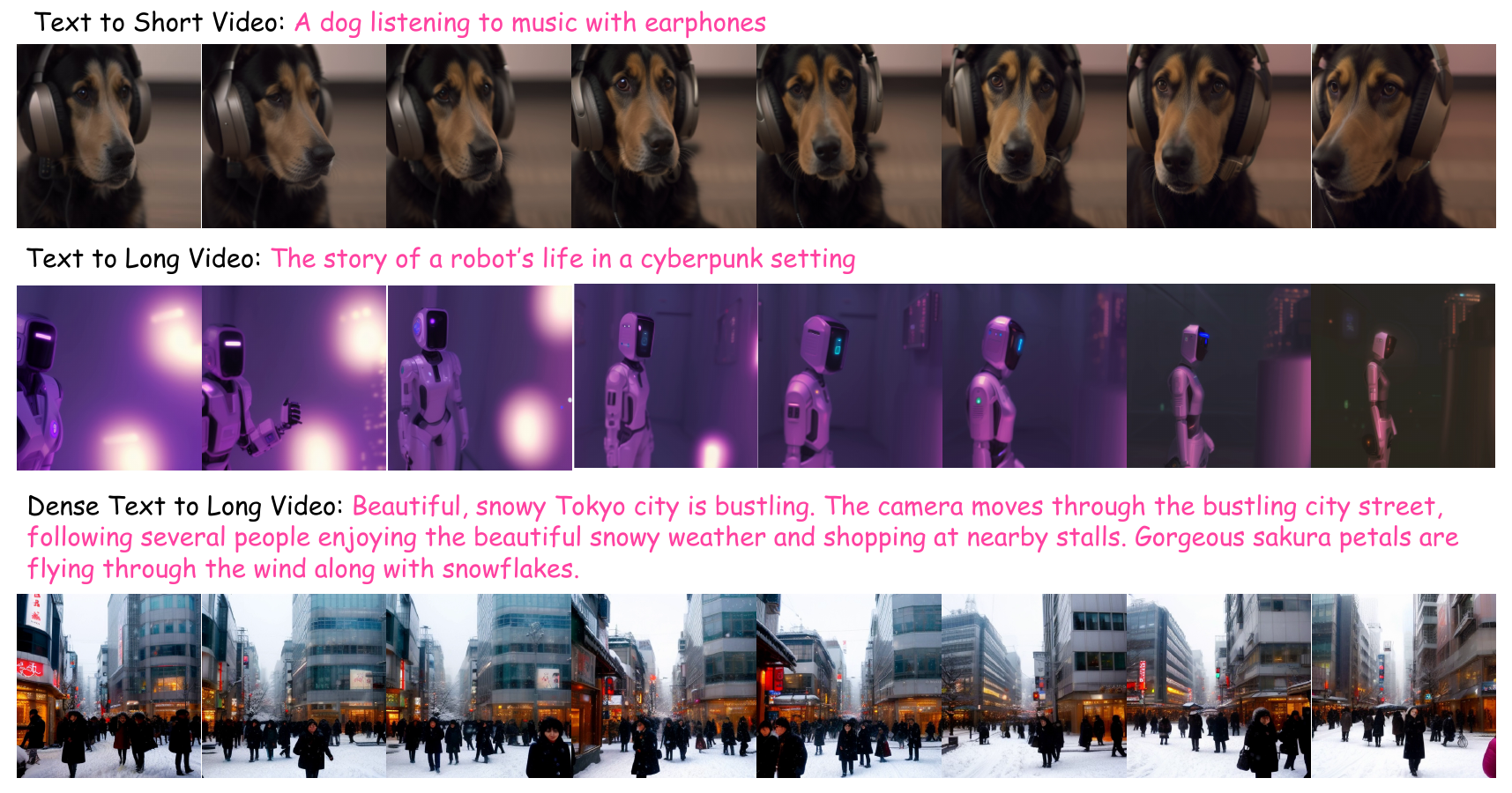}
    \caption{\textbf{Generated Videos from \Ours across Various Text-to-video Scenarios.} Our model demonstrates diversity and quality across various text-to-video tasks, including short video, long video, and dense text-to-long video generation. The results exhibit rich details, smooth transitions, and strong semantic alignment with input descriptions.
    }
    \vskip -0.3in
    \label{fig:vis_res}
\end{figure}

Fig.~\ref{fig:vis_res} illustrates the video frames generated by our model under various text-to-video generation scenarios.

\textbf{Text to Short Video.} In the top row of the figure, we show sample of short video generation. As shown in the figure, our approach has the capability to generate short videos with rich details and high fidelity while maintaining strong alignment with the given text descriptions.

\textbf{Text to Long Video.} The second row shows frames sampled from a long video generated by our model, conditioned on a concise text description. This sample demonstrate that our approach can generate long videos containing diverse content and larger dynamic changes compared to short video generation, while maintaining semantic alignment with the given text.

\textbf{Dense Text to Long Video.} Although not explicitly trained on dense captions, our model can effectively adapt to dense text video generation in a zero-shot manner. As illustrated in the last row of Fig.~\ref{fig:vis_res}, the generated long video depicts rich content that corresponds to the detailed descriptions, including multiple characters, weather, scenery, and building information. However, we observe that the generated images appear slightly blurry. We attribute this to the low resolution of our transformer's output, which may result in blurriness when generating highly detailed content.

\begin{figure}[htbp]
    \centering
    \includegraphics[width=1.0\textwidth]{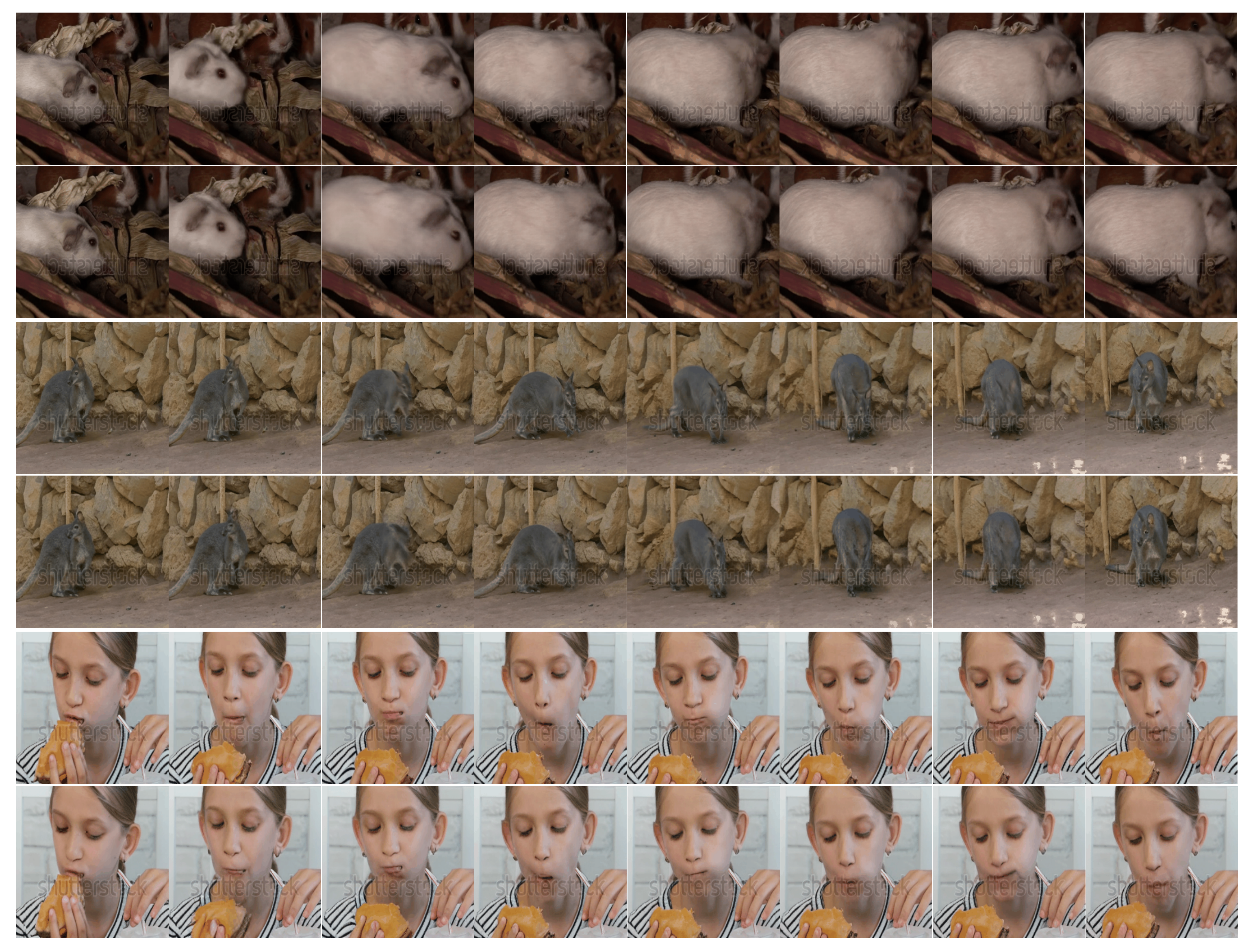}
    \caption{\textbf{Reconstructed Videos by Our Tokenizer.}  Each group represents a distinct video sequence, with the top row displaying the original frames and the bottom row presenting the corresponding reconstructions. Despite a high compression ratio of 256, our tokenizer effectively preserves fine details and natural, coherent motion in the reconstructed videos.
    }
    \label{fig:tokenizer}
\end{figure}

In Fig.~\ref{fig:tokenizer}, we also present a visualization of the videos reconstructed by our tokenizer. We use videos selected from the WebVid valiation dataset~\cite{Bain21} (not used for training). The original video frames are in the top row of each group, and the reconstructed videos of our tokenizer are shown in the bottom row. Despite achieving a high compression ratio of approximately 256 ($8\times8\times4$), our tokenizer effectively preserves the fine-grained details of the original frames, and also maintains natural and coherent motion along the temporal dimension.
\section{Conclusion and Discussions}
\label{conclusion}

In conclusion, we propose Loong, an autoregressive LLM-based video generation model that can generate minute-level long videos with consistent appearance, large motion dynamics, and natural scene transitions. We choose to model the text tokens and video tokens in a unified sequence, and overcome the challenges of long video training with the progressive short-to-long training scheme and loss re-weighting. Our experiments demonstrate the effectiveness of our approach in generating minute-level long videos. We hope our work can motivate research on long video generation and multimodal modeling in the future. 

\textbf{Border impact.} The model can be deployed to assist visual artists and film producers on video creation, enhancing their efficiency. It can also be deployed for entertainment purposes. On the other hand, it may be used for generating fake content and delivering misleading information. The community should be aware of the potential social impacts. It is necessary to develop techniques to detect and watermark the videos generated by machine learning models.

{
\bibliographystyle{ieee_fullname}

\bibliography{ref}
}

\end{document}